\documentclass{article}

\usepackage{arxiv}
\usepackage{microtype}
\usepackage{amsmath, amssymb, amsfonts}
\usepackage{booktabs, multicol, multirow}
\usepackage{algorithm2e}
\usepackage{graphicx} \graphicspath{ {./images/} }
\usepackage[hidelinks]{hyperref}

\title{Semi-Supervised Relational Contrastive Learning}

\author{
 Attiano Purpura-Pontoniere \\
  Computer Science Department\\
  University of California, Los Angeles\\
  \texttt{attiano@cs.ucla.edu} \\
   \And
  Demetri Terzopoulos \\
  Computer Science Department\\
  University of California, Los Angeles\\
  \texttt{dt@cs.ucla.edu}
  \And
  Adam Wang \\
  Radiology Department\\
  Stanford University\\
  \texttt{adamwang@stanford.edu} \\
   \And
  Abdullah-Al-Zubaer Imran \\
  Computer Science Department\\
  University of Kentucky\\
  \texttt{aimran@uky.edu}   
}

\begin{document}
\maketitle

\begin{abstract}
Disease diagnosis from medical images via supervised learning is usually dependent on tedious, error-prone, and costly image labeling by medical experts. Alternatively, semi-supervised learning and self-supervised learning offer effectiveness through the acquisition of valuable insights from readily available unlabeled images. We present \emph{Semi-Supervised Relational Contrastive Learning (SRCL)}, a novel semi-supervised learning model that leverages self-supervised contrastive loss and sample relation consistency for the more meaningful and effective exploitation of unlabeled data. Our experimentation with the SRCL model explores both pre-train/fine-tune and joint learning of the pretext (contrastive learning) and downstream (diagnostic classification) tasks. We validate against the ISIC 2018 Challenge benchmark skin lesion classification dataset and demonstrate the effectiveness of our semi-supervised method on varying amounts of labeled data. 
\end{abstract}


\section{Introduction}

When afforded full supervision for a target task, e.g., disease classification, convolutional neural networks (CNNs) can perform exceedingly well. However, the expertise required to produce the necessarily large, annotated training image datasets often makes supervised learning impractical in the medical domain. Self/semi-supervised learning approaches can be employed to alleviate the burden of labeling large datasets. Self-supervised learning exploits unlabeled data in order to produce a meaningful representation in an unsupervised manner \cite{doersch2017multitask}. Semi-supervised learning uses weak labels, or small labeled and larger unlabeled batches for concurrent supervised and unsupervised objectives \cite{chapelle2006semi}. Recently, contrastive self-supervised learning has shown promise \cite{liu2021self, wu2021federated, wang2020understanding, he2020momentum}. 

In contrastive learning, unlabeled data are leveraged to learn a generalized representation based on invariant perturbation. An embedding is created by transforming an input image and coercing the model to match the representations of the image and its transformation \cite{chen2020simple}. Relation consistency guides the predictions of a student model to match that of a teacher model under perturbation and noise injection, where a relation can be defined in terms of a connection or comparison between outputs of corresponding layers in the student and teacher models. The student model learns from supervised examples; i.e.,  it is taught like a student. The teacher model learns through knowledge transfer from the student model, with the weights of the student model serving to update the weights of the teacher model \cite{liu2020semi}.

Departing from existing self/semi-supervised techniques, we combine the orthogonal strength of Sample Relation Consistency (SRC) with supervised contrastive learning \cite{liu2020semi}. SRC forces each input image in a batch to maintain the same relation to itself and all other images in the batch, under perturbation and noise injection \cite{liu2020semi}. Supervised contrastive learning follows the contrastive learning paradigm, clustering different classes by forcing them to have similar representations in a down-sampled embedding space \cite{khosla2020supervised}. 

Within our novel Semi-Supervised Relational Contrastive Learning (SRCL) framework,
we explore the impact of label percentage with two different training processes: First, we investigate pre-training using supervised contrastive loss on a student model, then add the student model into the SRC architecture in conjunction with a mean teacher model. Second, we combine supervised contrastive loss and SRC to form a multi-task learning objective. We compare our models to the baseline of SRC and the baseline of complete supervision under cross entropy loss. We demonstrate that both pre-training with contrastive learning then fine-tuning with relational learning, as well as contrastive learning combined with relational learning, outperform the fully supervised model when the ratio of labeled training images is 50\% or less, and that the pre-training architecture consistently outperforms the SRC baseline. 

\section{Related Work}

Self-supervised and semi-supervised learning (SSL) have improved performance on downstream medical image analysis tasks such as classification and segmentation \cite{imran2020fully, jing2020self, imran2020self}. SSL models have also been shown to perform equally well on downstream tasks with only a subset of the data labeled. One example of self-supervised learning is semantic in-painting of cropped ultrasound images using conditional generative adversarial networks \cite{hu2020self}. 

An approach on which we focus is contrastive learning. SimCLR is the fundamental example, where each image in an input batch is transformed through data augmentations into a pair of images, resulting in a total of $2N$ images for a batch of $N$ images. Each transformed pair is passed through an encoder and a projection head that results in an embedding of both the original image and its transformation. The cosine similarity between the two embeddings of the image is maximized, while simultaneously minimizing the cosine similarity between those two embeddings (the positive set) and all other transformed image embeddings in the batch (the negative set) \cite{chen2020simple}. Since SimCLR, there have been a variety of additional developments in the contrastive learning framework \cite{azizi2021big, imran2022multimodal}, often increasing the size of the positive or negative sets. For example, recent work \cite{azizi2021big} added to the positive set distinct images that are different views of the same patient/condition. Another example is using metadata from MRI images to define positive pairs from 3D volumes as 2D slices that correspond to the same physical location; e.g., the same part of the lungs, across different patients \cite{chaitnaya2020contrastive}. A third contrastive based example is the use of  contrastive learning across modalities (lateral and frontal scouts) to estimate patient-specific CT radiation doses \cite{imran2022multimodal}.

A similar example, which is used as a building block in our work, is supervised contrastive loss. For a given image, it adds to the contrastive positive set all other images in the batch with the same label \cite{khosla2020supervised}. The Leave-One-Out contrastive learning framework \cite{xiao2020what} prescribes an additional embedding space for each augmentation, and the projection head that defines a given embedding is taught to be sensitive to a specific augmentation, yielding improved robustness in cases where valuable cues from the augmentation are meaningful in the downstream task; e.g., color for bird classification. 

Relation consistency is another semi-supervised approach that seeks to maintain the consistency between predictions of separate models for unlabeled and labeled data \cite{DBLP:journals/corr/LaineA16/pi_model}. Sample relation consistency, another building block for this paper, goes a step further to maintain the same relationship between all samples in a batch by minimizing the $L^2$ norm between the activation matrices of separate teacher and student models \cite{liu2020semi}. Another example of the use of consistency-based methods to alleviate the labeling necessity is  consistency-based learning to train highly accurate models on point-based instead of region-based segmentation of CT images \cite{laradji2020weakly}. In a similar vein, contrastive learning has been applied to capture feature representations from large lung datasets by adopting a standard classification network, demonstrating superior performance in the diagnosis of COVID-19 from chest CT images in two datasets \cite{CHEN2021Momentum}.

\begin{figure*}
     \centering
     \includegraphics[width=\linewidth]{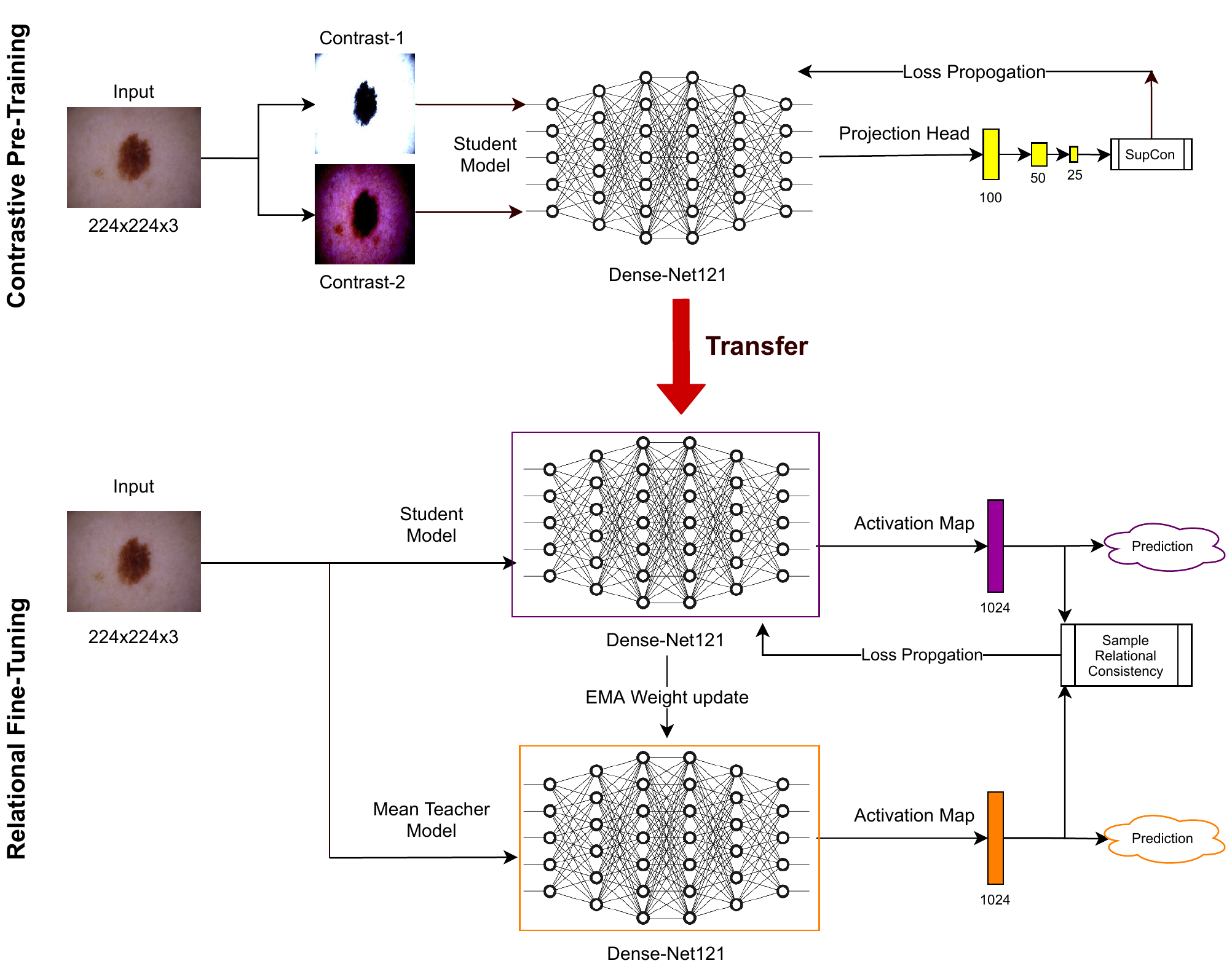}
    \caption{The Semi-Supervised Relational Contrastive Learning (SRCL) model. The backbone network is based on Dense-Net121.  We first pre-train using supervised contrastive loss (top). In the fine-tuning stage (bottom) for the downstream classification task, we employ sample relation consistency with the mean teacher and student model.}
    \label{fig:pretrain_model}
\end{figure*}

\begin{algorithm}[t]
\caption{SRCL Training Algorithm: Pre-Training}
\label{alg:pre}
\KwIn{batch size $N$, subset of batch labeled $L$, constant $\tau$, model $f_\text{student}$, contrastive projection head $g$, set of augmentations $T$, pre-training epochs $E_\text{pre}$}
\KwOut{update networks $f_\text{student}$ and $g$ to minimize loss $\mathcal{L}$}
\For{each $E_\text{pre}$ epoch}{
    \For{Sampled batch of size $N$ with subset of labels $\{y_i\}_{i=1}^L$}{
            \For{$k=1$  \KwTo $N$}{
                draw two augmentation functions $t \sim T$, $t' \sim T$;\\
                
                $\hat{x}_{2k-1}=t(x_{k})$; \CommentSty{//first transformed image}\\
                $\hat{h}_{2k-1}=f_\text{student}(\hat{x}_{2k-1})$; \CommentSty{//student activation}\\
                $z_{2k-1}=g(h_{2k-1})$; \CommentSty{//contrastive projection}\\
                $\hat{x}_{2k}=t(x_{k})$; \CommentSty{//second transformed image}\\
                $\hat{h}_{2k}=f_\text{student}(\hat{x}_{2k})$; \CommentSty{//student activation map}\\
                $z_{2k}=g(h_{2k})$; \CommentSty{//contrastive projection}\\
            $\mathcal{L} = \frac{1}{2N} \sum^N_{k=1} [\mathcal{L}_\text{SupCon}(z_{2k},z_{2k-1}, y_i) + \mathcal{L}_\text{SupCon}(z_{2k-1},z_{2k}, y_i)]$;
        }
    }
}
\end{algorithm}

\begin{algorithm}[t]
\caption{SRCL Training Algorithm: Fine-Tuning}
\label{alg:fine}
\KwIn{batch size $N$, subset of batch labeled $L$, constant $\tau$, model $f_\text{teacher}$, relation matrix transformation $R$, classification network $C$, set of augmentations $T$, warm-up epochs $E_\text{warm}$, training epochs $E_\text{train}$, transfer encoder network $f_\text{student}$ from pre-training
}
\KwOut{encoder network $f_\text{student}$ and classification network $C$}
\For{each $E_\text{train}$ epoch}{
    \For{Sampled batch of size $N$ with subset of labels $\{y_i\}_{i=1}^L$}{
        \For{$k=1$ \KwTo $N$}{
            $\hat{h}_{k}=f_\text{student}(\hat{x}_{k})$; \CommentSty{//student activation}\\
            $h_{k}=f_\text{teacher}(x_{k})$; \CommentSty{//teacher activation}\\
            \If{$k \in \{1,...,L\}$}{
                $\hat{\text{pred}}_{k}=C(\hat{h}_{k})$; \CommentSty{//student prediction}\\
                $\text{pred}_{k}=C(h_{k})$; \CommentSty{//teacher prediction}\\
                $\mathcal{L} \mathrel{+}= \mathcal{L}_\text{MSE}(\hat{\text{MSE}}_{k},\text{pred}_{k})$;\\
            }
        }
    }
    $\hat{G} = \hat{\mathbf{h}} \hat{\mathbf{h}}^T$; \CommentSty{//student model Gram matrix}\\
    $G = \mathbf{h} \mathbf{h}^T$; \CommentSty{//teacher model Gram matrix}\\
    \If{$\text{epoch} \geq \text{Epoch}_\text{warm}$}{
        $\mathcal{L} \mathrel{+}= \mathcal{L}_\text{SRC}(R(\hat{G}),R(G))$;\\
    }
    update networks $f_\text{student}$ and $C$ to minimize $\mathcal{L}$;\\
    update network $f_\text{teacher}$ as exponential moving average of $f_\text{student}$:
    $f_{\text{teacher}}^e = \alpha f_{\text{teacher}}^{e-1} + (1-\alpha) f_{\text{student}}^e$
}
\end{algorithm}

\section{The SRCL Model}

Fig.~\ref{fig:pretrain_model} illustrates our Semi-Supervised Relational Contrastive Learning (SRCL) model. 

We use a Dense-Net121 architecture as the backbone for both the pre-training and fine-tuning stages of the model. Like other relation consistency architectures \cite{tarvainen2017mean}, we use a pair of models, a student and a teacher model. At first, we pre-train the student model using supervised contrastive loss for $E_\text{pre}$ epochs. We then perform SRC to fine-tune the model for $E_\text{down}$ epochs. In every epoch, the weights of the teacher model update as the exponential moving average of the weights of the student model. However, we begin to update the teacher model only after a sufficient number $E_\text{warm}$ of ``warm-up epochs'', such that the student model learns a non-random embedding of the input data. 

Traditional relation consistency enforces the individual sample predictions of the student and teacher modeld to match under perturbation. SRC and SRCL both use the entire sample batch to maintain consistent embeddings between the student and teacher models across the sample batch; i.e., each sample maintains the same relation to every other sample in the batch. The student model learns from supervised examples, and the teacher model is not updated until the student model has gone through 20 epochs of supervised training.
We feed unlabeled examples into both the student and teacher models to encourage relation consistency between predictions over the unlabeled data. When applying contrastive learning, SRCL uses any labels available for a given batch of images, which provides a larger set of positive pairs against which to contrast. 

Algorithms~\ref{alg:pre} and~\ref{alg:fine} illustrate the overall training procedure of our SRCL model during pre-training and fine-tuning, respectively. The training objective of the SRCL model includes a sample relational consistency loss and a supervised contrastive learning loss. The sample relational consistency loss $\mathcal{L}_\text{SRC}$ is 
\begin{equation}
    \mathcal{L}_\text{SRC} = \sum_N \frac{1}{N} \| R_\text{student} - R_\text{teacher}\|^2
\end{equation}
where $N$ is the number of images in the input batch, and $R_\text{student}$ and $R_\text{teacher}$ are the sample relation matrices formed by taking the $L^2$ row-wise norm of Gram Matrix $G = A A^T$, where the activation maps $A \in \mathbb{R}^{NxH*W*C}$ are formed from the concatenation of the height, width, and channels of the encoding of a given batch of input images \cite{liu2021self}.

We calculate the supervised contrastive loss as  
\begin{equation}
    \mathcal{L}_\text{SupCon} = \sum_{i \epsilon I} \frac{1}{|P(i)|} \sum_{p \epsilon P(i)} \log{\frac{e^{\frac{z_i \cdot z_a}{\tau}}}{\sum_{\alpha \epsilon A(i)} e^{\frac{z_i \cdot z_a}{\tau}}}},
\end{equation}
where $i \in I \equiv \{1..2N\}$ is contrastive image $i$ created in pairs from an input batch of $N$ images, $A(i) \equiv I-i$ is the set of all transformed images excluding contrastive image $i$, and $P(i) \equiv \{p \in A(i): y_p = y_i\}$ is the set of all positive images for contrastive image $i$; i.e., all transformed images in this batch that have the same label as image $i$. Without the labels, this loss function degenerates to SimCLR \cite{chen2020simple}. The projection of the activation map for a given contrastive image $i$ transformed from a corresponding original image $x$ is $z_i = \text{Proj}(\text{Enc}(\hat{x_i}))$. Finally, $\tau$ is a temperature scaling constant that acts as a hyperparameter to adjust the similarity importance between positive pairs~\cite{khosla2020supervised}.

The mean squared error loss used for standard supervision is
\begin{equation}
    \mathcal{L}_\text{MSE} = \frac{1}{N}\sum_N (Y_i - \hat{Y}_i)^2,
\end{equation}
where $Y_i$ is the prediction for a given sample and $\hat{Y}_i$ is the ground truth label \cite{berger1985statistical}.

\section{Experimental Evaluation}

\subsection{Implementation Details}

\begin{figure}
    \centering
    \includegraphics[width=0.8\linewidth, trim={2.5cm 1cm 3cm 1.3cm}, clip]{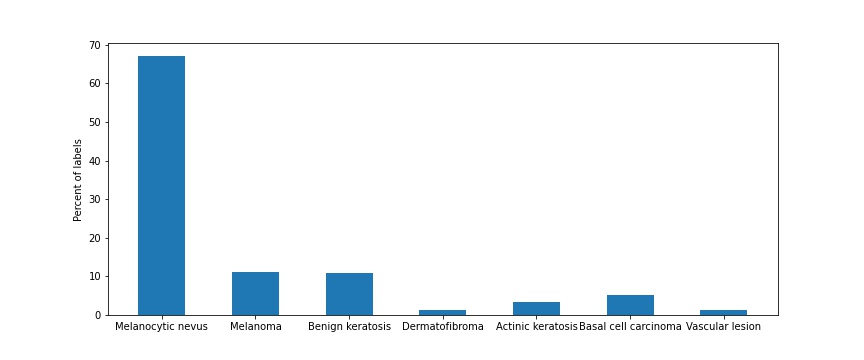}
    \caption{Label distribution across the 7 different skin lesion condition classes present in the ISIC 2018 dataset.}
    \label{fig:label_distro}
\end{figure}

\paragraph{\bf Data:} We used 10,015 dermoscopy images with 7 common skin lesion conditions from the ISIC 2018 skin lesion analysis dataset \cite{codella2019skin}. The dataset was partitioned into train (7,000), validation (1,000), and test (2,000) sets. We resized all images to $224\times224$ pixels and the contrastive images were transformed through a random crop, color jitter, grey-scaling, and normalization based on a statistic obtained from ImageNet \cite{russakovsky2015imagenet}.
Fig.~\ref{fig:label_distro} shows the label distribution for the 7 skin lesion conditions considered in the dataset: Melanocytic Nevus, Melanoma, Benign Keratosis, Dermatofibroma, Actinic Keratosis, Basal Cell Carcinoma, and Vascular lesions. Clearly, the vast majority of labels are skewed towards Melanocytic Nevus, which is roughly 70\% of the dataset, with the remainder mostly spread between Melanoma and Benign Keratosis, both of which comprise roughly 10\% of the dataset. The training, test, and validation sets exhibit the same label distribution skew. 

\paragraph{\bf Model Training:} We passed the images through the DenseNet121 backbone with pretrained weights from ImageNet. Mostly, we ran the models using Google Colab and Colab Pro, which provides a Tesla P4, T4, P100, and occasionally a VX100 GPU. Due to the inconsistency of the GPUs, we performed all the experiments with a ``lowest common denominator'' batch size of 20. A larger batch size would provide better results as contrastive learning performs better with larger batch sizes by creating larger positive and negative sets \cite{chen2020simple}. 
We set $\tau = 0.1$ through an exhaustive search procedure. In our experiments, we trained with $E_\text{pre}=E_\text{down}=100$, $E_\text{warm}=20$, and chose the best model based on the validation performance.  Through ablation studies, we found the best projection head size to be a series of three fully-connected layers of sizes 100, 50, and finally 25.

\subsection{Results and Discussion}

\begin{table}[t]
    \centering
    \setlength{\tabcolsep}{2pt}
    \caption{Performance comparison of our SRCL models against the baselines across varying labeled data percentages. The highest performance figures are in boldface.}
    \smallskip
    \label{table:new}
    \begin{tabular}{@{} lc cc cccc @{}}
    \toprule
    \multirow{2}{*}{Model} & \phantom{a} & \multirow{2}{*}{\%Labeled} & \phantom{a} & \multicolumn{4}{c}{Performance Metrics}\\
    \cmidrule{5-8}
    & & & & AUROC & Accuracy & Sensitivity & Specificity\\
    \midrule
    \multirow{7}{*}{\rotatebox{30}{Supervised}} 
    && 5\% && 0.765 & 0.863 & \textbf{0.349} & 0.902
    \smallskip\\
    && 10\% && 0.864 & 0.888 & \textbf{0.487} & \textbf{0.929}
    \smallskip\\
    && 20\% && 0.878 & 0.909 & \textbf{0.635} & \textbf{0.941}
    \smallskip\\
    && 30\% && 0.881 & 0.918 & \textbf{0.654} & \textbf{0.944}
    \smallskip\\
    && 40\% && 0.936 & 0.921 & \textbf{0.757} & \textbf{0.949}
    \smallskip\\
    && 50\% && 0.940 & 0.934 & \textbf{0.784} & \textbf{0.951}
    \smallskip\\
    && 100\% && \textbf{0.959} & \textbf{0.950} & \textbf{0.805} & \textbf{0.955}
    \smallskip\\
    \midrule
    \multirow{6}{*}{\rotatebox{30}{SRC}} 
    && 5\% && 0.798 & 0.913 & 0.306 & 0.898
    \smallskip\\
    && 10\% && 0.878 & 0.924 & 0.438 & 0.904
    \smallskip\\
    && 20\% && 0.896 & 0.932 & 0.599 & 0.915
    \smallskip\\
    && 30\% && 0.915 & 0.935 & 0.580 & 0.922
    \smallskip\\
    && 40\% && 0.937 & 0.938 & 0.631 & 0.926
    \smallskip\\
    && 50\% && 0.941 & 0.944 & 0.723 & 0.930
    \smallskip\\

    \midrule
    \multirow{6}{*}{\rotatebox{30}{SRCL--Joint}} 
    && 5\% && \textbf{0.799} & 0.910 & 0.301  & 0.902
    \smallskip\\
    && 10\% && 0.865 & 0.922 & 0.364 & 0.899
    \smallskip\\
    && 20\% && 0.921 & 0.934 & 0.541 & 0.916
    \smallskip\\
    && 30\% && 0.927 & 0.923 & 0.549 & 0.915
    \smallskip\\
    && 40\% && 0.931 & 0.924 & 0.631 & 0.920
    \smallskip\\
    && 50\% && 0.918 & 0.930 & 0.643 & 0.914
    \smallskip\\
    \midrule
    \multirow{6}{*}{\rotatebox{30}{SRCL}} 
    && 5\% && 0.779 & \textbf{0.915} & 0.338 & \textbf{0.905}
    \smallskip\\
    && 10\% && \textbf{0.886} & \textbf{0.926} & 0.400 & 0.902
    \smallskip\\
    && 20\% && \textbf{0.925} & \textbf{0.936} & 0.586 & 0.919
    \smallskip\\
    && 30\% && \textbf{0.934} & \textbf{0.938} & 0.633 & 0.927
    \smallskip\\
    && 40\% && \textbf{0.948} & \textbf{0.943} & 0.676 & 0.933
    \smallskip\\
    && 50\% && \textbf{0.950} & \textbf{0.946} & 0.720 & 0.931
    \smallskip\\
    \bottomrule
  \end{tabular}
\end{table}

Table~\ref{table:new} reports our results for 5\%, 10\%, 20\%, 30\%, 40\%, and 50\% labels. We also include the fully supervised baseline with 100\% labels to serve as a performance upper bound. We report the Area Under the Receiver Operating Characteristic Curve (AUROC), the accuracy, the sensitivity, and the specificity. We also report metrics for the fully supervised upper bound, the baseline SRC, our pre-training SRCL, and finally our combined training SRCL-Joint. We report metrics on the testing set. For fairness, all models used the same hyper-parameter values specified above.

Table~\ref{table:ssl} reports accuracy results for our SRCL model in comparison to various other SSL methods. We also include the fully supervised upper bound with 100\% labels. We compare against the accuracies reported in the SSL literature on skin leison classification. For a fair comparison, all models used the same ISIC 2018 dataset, and the same quantity of labels from that dataset.

\begin{table}[t]
    \centering
    \setlength{\tabcolsep}{2pt}
    \caption{Performance comparison of the baseline, our SRCL model, and the other SSL methods using 20\% labeled data, with the highest metric in boldface for the best performing model and underlined for the second best performing model.}
    \smallskip
    \label{table:ssl}
    \begin{tabular}{lc c c}
        \toprule
         Model & \phantom{a}  & \%Labeled & Accuracy \\
         \midrule
         Supervised (upper-bound) && 100 & 0.950 \\
         \midrule
         Supervised Baseline && 20  & 0.909 \\
         MT \cite{liu2021semi} && 20 & 0.917 \\
         SE-MT \cite{kitada2018skin} && 20 & 0.872\\
         DDM \cite{liu2021semi}&& 20 & \textbf{0.940}\\
         FedIRM \cite{liu2021federated} && 20  & 0.929 \\ 
         SRCL (ours) && 20 & \underline{0.936}\\
         \bottomrule
    \end{tabular}
\end{table}

\begin{figure}[t]
    \centering
    \includegraphics[width=0.5\linewidth, trim={0.25cm 0.2cm 1cm 1cm}, clip]{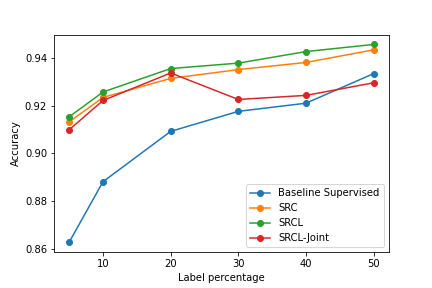}
    \caption{Accuracy vs labeled data percentage for the supervised baseline, state-of-the-art SRC, and our SRCL models.}
    \label{fig:acc_comparison}
\end{figure}

\begin{figure*}
    \centering
    \includegraphics[width=0.4\linewidth]{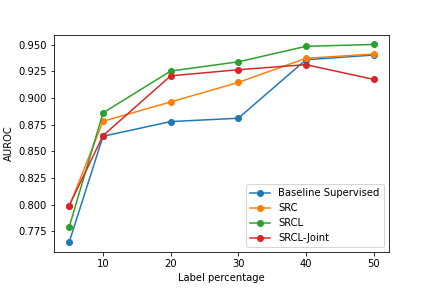}
    \hfill
    \includegraphics[width=0.59\linewidth, trim={2.75cm 0.5cm 3cm 1.4cm}, clip]{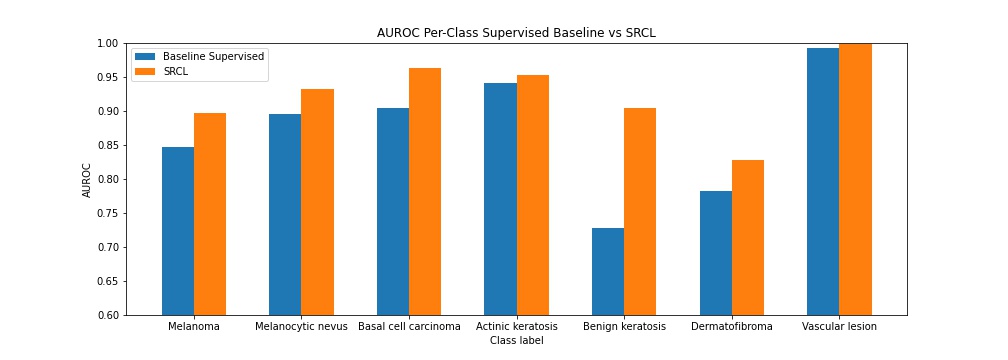}
    \caption{The superior performance of our SRCL model. (left) Consistent AUROC performance improvement with an increase in labeled data by our SRCL model over the fully supervised baseline and state-of-the-art SRC models. (right) Per-class AUROC performance for the fully supervised baseline and our SRCL model with 20\% labeled data.}
    \label{fig:label_results}
\end{figure*} 

Fig.~\ref{fig:acc_comparison} reports the accuracies of the baseline and our models across various label percentages, from 5\% to 50\% labels. Clearly, our best performing SRCL model consistently outperforms the baselines.

We experimented with our SRCL and SRCL-Joint 
models and compared against a supervised baseline (Dense-Net121) and the state-of-the-art SRC \cite{liu2020semi} methods. As shown in Fig.~\ref{fig:label_results}, both the SRCL and SRCL-Joint model AUROC scores consistently improve with an increase in the label percentages. Even at 5\% labeled data, SRCL-Joint outperforms the other models, affirming the significant gain in learning derived from unlabeled data in joint training. The two-step SRCL model performs better than the end-to-end SRCL-Joint model at higher percentages of labeled data. This reveals that SRCL is able to initialize the student model with a better encoding function, enabling the teacher model to be initialized with better starting weights for consistency. The superiority of SRCL over SRC may be explained by the fact that relational learning must train the student model to learn an accurate representation to transfer to the teacher model. With fewer labels, the embedding learned by the student model is not precise enough compared to pre-training in SRCL via contrastive learning. The consistent improvement with either of our models over the baselines demonstrates the effectiveness of combining supervised contrastive learning with relation consistency learning. 

Our SRCL model achieves an overall accuracy of 0.946, which is comparable to the 0.950 accuracy of the fully supervised baseline, while reducing the annotation cost by 50\%. 

Moreover, the class-wise AUROC comparison at 20\% labels (Fig.~\ref{fig:label_results}) reveals the robustness of our SRCL model in classifying the dermoscopy images. On vascular lesions, the results are very close, since vascular lesions are very distinct from the surrounding skin. They are large, red, and burgundy colored, and bulbous in comparison to the neighboring skin, and they appear similar across cases. As such, vascular lesions are an easy class to identify after a basic vascular lesion representation has been learned.

\begin{figure*}
\centering
\resizebox{\linewidth}{!}{
\begin{tabular}{c c c}
\includegraphics[width=0.33\linewidth, trim={1cm 2cm 8cm 1cm}, clip]{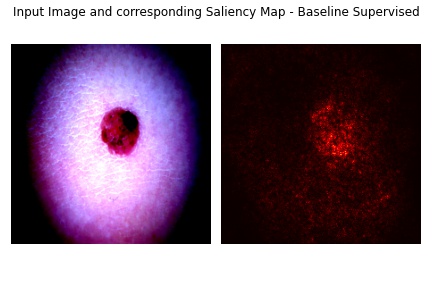} &
\includegraphics[width=0.33\linewidth, trim={8cm 2cm 1cm 1cm}, clip]{images/baseline_0034286_saliency.jpg}&
\includegraphics[width=0.33\linewidth, trim={8cm 1cm 1cm 1cm}, clip]{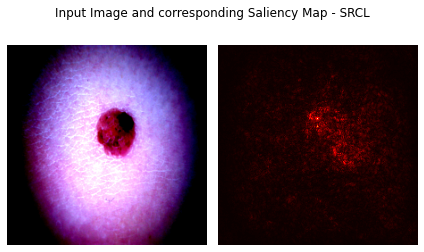} \\

\includegraphics[width=0.33\linewidth, trim={3cm 1cm 3cm 0cm},clip]{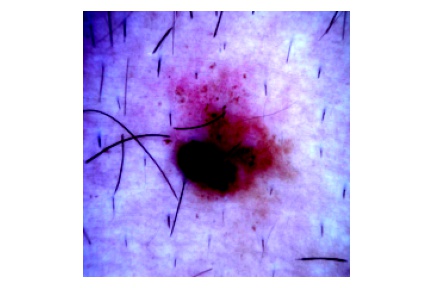} &
\includegraphics[width=0.33\linewidth, trim={3cm 1cm 3cm 0cm}, clip]{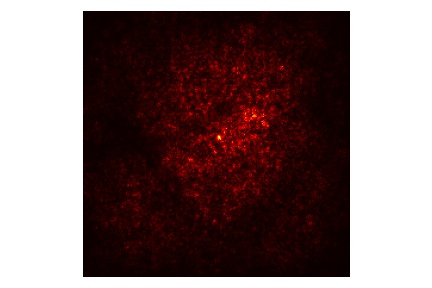} &
\includegraphics[width=0.33\linewidth, trim={3cm 1cm 3cm 0cm}, clip]{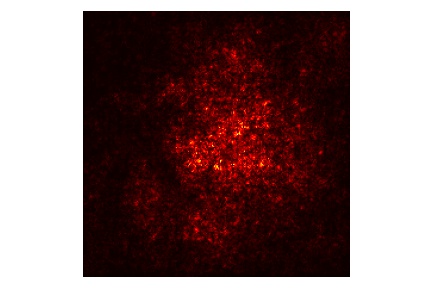} \\

\includegraphics[width=0.33\linewidth, trim={3cm 1cm 3cm 0cm},clip]{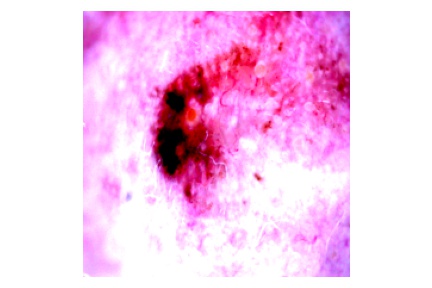} &
\includegraphics[width=0.33\linewidth, trim={3cm 1cm 3cm 0cm}, clip]{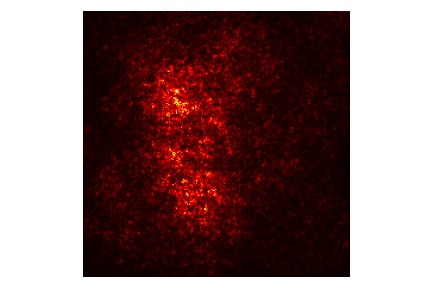} &
\includegraphics[width=0.33\linewidth, trim={3cm 1cm 3cm 0cm}, clip]{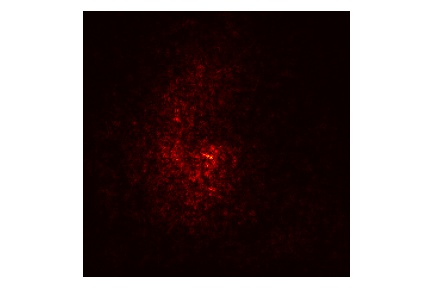} \\

\includegraphics[width=0.33\linewidth, trim={3cm 1cm 3cm 0cm},clip]{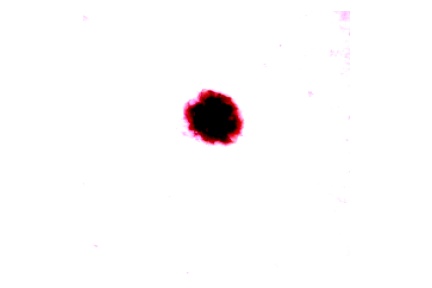} &
\includegraphics[width=0.33\linewidth, trim={3cm 1cm 3cm 0cm}, clip]{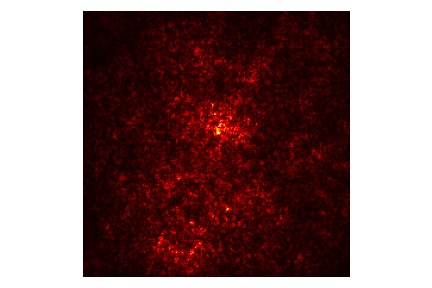} &
\includegraphics[width=0.33\linewidth, trim={3cm 1cm 3cm 0cm}, clip]{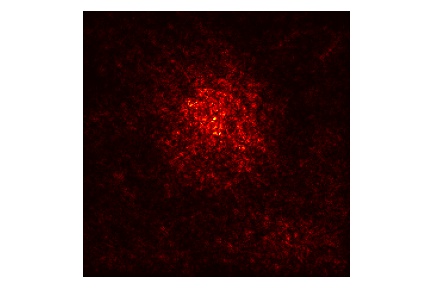} \\
{\large Input} & {\large Supervised} & {\large SRCL}
\end{tabular}
}
\caption{Saliency visualization at 20\% labeled data demonstrates the superiority of our SRCL model over the baseline supervised model in attending to important pixels in the image.}
\label{fig:saliency_maps}
\end{figure*}

Saliency map visualization further reveals the superiority of our SRCL models. The saliency map visualizes the gradient of the model output when performing inference on an input image.
Fig.~\ref{fig:saliency_maps} shows example input images and corresponding saliency maps for the fully supervised baseline and the corresponding saliency maps for our best performing SRCL model.  Both models are evaluated when 20\% of the data is labeled. Clearly, the models are attending to the relevant regions of the input image; e.g., the pixels that contain the skin lesion, as the gradients for those pixels are the highest. Our model attends more to the relevant regions than the supervised baseline, as the pixels not corresponding to the lesion in the saliency map are darker, indicating lower gradient components.

Our experimentation reveals a clear performance trend---with up to 50\% unlabeled data, our SRCL model performs comparably to the supervised baseline model. The scope of our experiments was to improve the performance of our models with limited labeled data; therefore, we performed more thorough experimentation with less than half the dataset labeled. SRCL-Joint shows a drop in performance from 50\% to 40\%  labeled data, as well as the lowest overall performance, which likely occurs because it has competing multi-task objectives that are not completely orthogonal. 

We hypothesize that these tasks are not completely orthogonal as contrastive learning tries to train the model to cluster positive examples and de-cluster negative examples, while relational learning tries to maintain the same clustering present in the original batch under perturbation. In other words, relational learning tries to hold things fixed in the embedding space under perturbation while contrastive learning tries to push things together and apart in the embedding space under perturbation. They are potentially competing objectives in the embedding space. It would likely improve performance to train a multi-task semi-supervised model to perform tasks either on disparate embedding spaces (e.g., natural language and image data), or that are complementary (detection and classification), or that have orthogonal effects in the embedding space.
    
\section{Conclusions}

We have demonstrated that contrastive learning and relational learning, when combined, can improve classification performance. Validating against the ISIC 2018 dataset on 7 skin lesion classifications, our proposed Semi-Supervised Relational Contrastive Learning (SRCL) framework outperforms the fully-supervised as well as the Sample Relation Consistency (SRC) methods considered. 

Future development will include adding more sophisticated augmentations, as well as more advanced contrastive learning strategies. Furthermore, additional imperatives could be added to the relational loss function in order to enforce consistency at intermediate layers so as to further regularize the features across the student and teacher models. Additional future work may include improving the contrastive learning aspect by exploring different advanced augmentations, such as  Mixup \cite{zhang2017mixup} or Cutmix \cite{yun2019cutmix}, and creating a projection head that varies for each specific augmentation; e.g., leaving one of the unique augmentations out for each projection head, which has been demonstrated to retain variance to signals that contain valuable information, such as color \cite{xiao2020what}.

\bibliographystyle{acm}
\bibliography{main}

\begin{thebibliography}{10}

\bibitem{azizi2021big}
{\sc Azizi, S., Mustafa, B., Ryan, F., Beaver, Z., Freyberg, J., Deaton, J.,
  Loh, A., Karthikesalingam, A., Kornblith, S., Chen, T., Natarajan, V., and
  Norouzi, M.}
\newblock Big self-supervised models advance medical image classification.
\newblock {\em arXiv preprint arXiv:2101.05224\/} (2021).

\bibitem{berger1985statistical}
{\sc Berger, J.~O.}
\newblock {\em Statistical Decision Theory and {Bayesian} Analysis, 2nd
  Edition}.
\newblock Springer Series in Statistics. Springer-Verlag, 1985.

\bibitem{chaitnaya2020contrastive}
{\sc Chaitanya, K., Erdil, E., Karani, N., and Konukoglu, E.}
\newblock Contrastive learning of global and local features for medical image
  segmentation with limited annotations.
\newblock {\em CoRR abs/2006.10511\/} (2020).

\bibitem{chapelle2006semi}
{\sc Chapelle, O., Sch{\"o}lkopf, B., and Zien, A.}, Eds.
\newblock {\em Semi-Supervised Learning}.
\newblock MIT Press, Cambridge, MA, 2006.

\bibitem{chen2020simple}
{\sc Chen, T., Kornblith, S., Norouzi, M., and Hinton, G.}
\newblock A simple framework for contrastive learning of visual
  representations.
\newblock In {\em International Conference on Machine Learning\/} (2020),
  pp.~1597--1607.

\bibitem{CHEN2021Momentum}
{\sc Chen, X., Yao, L., Zhou, T., Dong, J., and Zhang, Y.}
\newblock Momentum contrastive learning for few-shot {COVID-19} diagnosis from
  chest {CT} images.
\newblock {\em Pattern Recognition 113\/} (2021), 107826.

\bibitem{codella2019skin}
{\sc Codella, N., Rotemberg, V., Tschandl, P., Celebi, M.~E., Dusza, S.,
  Gutman, D., Helba, B., Kalloo, A., Liopyris, K., Marchetti, M., et~al.}
\newblock Skin lesion analysis toward melanoma detection 2018: A challenge
  hosted by the international skin imaging collaboration ({ISIC}).
\newblock {\em arXiv preprint arXiv:1902.03368\/} (2019).

\bibitem{doersch2017multitask}
{\sc Doersch, C., and Zisserman, A.}
\newblock Multi-task self-supervised visual learning.
\newblock In {\em 2017 IEEE International Conference on Computer Vision
  (ICCV)\/} (2017), pp.~2070--2079.

\bibitem{he2020momentum}
{\sc He, K., Fan, H., Wu, Y., Xie, S., and Girshick, R.}
\newblock Momentum contrast for unsupervised visual representation learning.
\newblock In {\em Proceedings of the IEEE/CVF Conference on Computer Vision and
  Pattern Recognition\/} (2020), pp.~9729--9738.

\bibitem{hu2020self}
{\sc Hu, S.-Y., Wang, S., Weng, W.-H., Wang, J., Wang, X., Ozturk, A., Li, Q.,
  Kumar, V., and Samir, A.~E.}
\newblock Self-supervised pretraining with {DICOM} metadata in ultrasound
  imaging.
\newblock In {\em Proceedings of the 5th Machine Learning for Healthcare
  Conference\/} (2020), pp.~732--749.

\bibitem{imran2020fully}
{\sc Imran, A.-A.-Z.}
\newblock {\em From Fully-Supervised, Single-Task to Scarcely-Supervised,
  Multi-Task Deep Learning for Medical Image Analysis}.
\newblock PhD thesis, Computer Science Department, University of California,
  Los Angeles, 2020.

\bibitem{imran2020self}
{\sc Imran, A.-A.-Z., Huang, C., Tang, H., Fan, W., Xiao, Y., Hao, D., Qian,
  Z., and Terzopoulos, D.}
\newblock Self-supervised, semi-supervised, multi-context learning for the
  combined classification and segmentation of medical images.
\newblock In {\em Proceedings of the AAAI Conference on Artificial
  Intelligence\/} (2020), vol.~34, pp.~13815--13816.

\bibitem{imran2022multimodal}
{\sc Imran, A.-A.-Z., Wang, S., Pal, D., Dutta, S., Zucker, E., and Wang, A.}
\newblock Multimodal contrastive learning for prospective personalized
  estimation of {CT} organ dose.
\newblock In {\em International Conference on Medical Image Computing and
  Computer-Assisted Intervention\/} (2022), pp.~634--643.

\bibitem{jing2020self}
{\sc Jing, L., and Tian, Y.}
\newblock Self-supervised visual feature learning with deep neural networks: A
  survey.
\newblock {\em IEEE Transactions on Pattern Analysis and Machine Intelligence
  43}, 11 (2020), 4037--4058.

\bibitem{khosla2020supervised}
{\sc Khosla, P., Teterwak, P., Wang, C., Sarna, A., Tian, Y., Isola, P.,
  Maschinot, A., Liu, C., and Krishnan, D.}
\newblock Supervised contrastive learning.
\newblock {\em arXiv preprint arXiv:2004.11362\/} (2020).

\bibitem{kitada2018skin}
{\sc Kitada, S., and Iyatomi, H.}
\newblock Skin lesion classification with ensemble of squeeze-and-excitation
  networks and semi-supervised learning.
\newblock {\em CoRR abs/1809.02568\/} (2018).

\bibitem{DBLP:journals/corr/LaineA16/pi_model}
{\sc Laine, S., and Aila, T.}
\newblock Temporal ensembling for semi-supervised learning.
\newblock {\em CoRR abs/1610.02242\/} (2016).

\bibitem{laradji2020weakly}
{\sc Laradji, I., Rodriguez, P., Mañas, O., Lensink, K., Law, M., Kurzman, L.,
  Parker, W., Vazquez, D., and Nowrouzezahrai, D.}
\newblock A weakly supervised consistency-based learning method for {COVID-19}
  segmentation in {CT} images.
\newblock {\em arXiv preprint arXiv:2007.02180\/} (2020).

\bibitem{liu2021semi}
{\sc Liu, P., and Zheng, G.}
\newblock Semi-supervised learning regularized by adversarial perturbation and
  diversity maximization.
\newblock In {\em Machine Learning in Medical Imaging\/} (Cham, 2021), C.~Lian,
  X.~Cao, I.~Rekik, X.~Xu, and P.~Yan, Eds., Springer International Publishing,
  pp.~199--208.

\bibitem{liu2021federated}
{\sc Liu, Q., Yang, H., Dou, Q., and Heng, P.}
\newblock Federated semi-supervised medical image classification via
  inter-client relation matching.
\newblock {\em CoRR abs/2106.08600\/} (2021).

\bibitem{liu2020semi}
{\sc Liu, Q., Yu, L., Luo, L., Dou, Q., and Heng, P.~A.}
\newblock Semi-supervised medical image classification with relation-driven
  self-ensembling model.
\newblock {\em IEEE Transactions on Medical Imaging 39}, 11 (2020), 3429--3440.

\bibitem{liu2021self}
{\sc Liu, X., Zhang, F., Hou, Z., Mian, L., Wang, Z., Zhang, J., and Tang, J.}
\newblock Self-supervised learning: Generative or contrastive.
\newblock {\em IEEE Transactions on Knowledge and Data Engineering 35\/}
  (2021), 857--876.

\bibitem{russakovsky2015imagenet}
{\sc Russakovsky, O., Deng, J., Su, H., Krause, J., Satheesh, S., Ma, S.,
  Huang, Z., Karpathy, A., Khosla, A., Bernstein, M., et~al.}
\newblock Imagenet large scale visual recognition challenge.
\newblock {\em International Journal of Computer Vision 115}, 3 (2015),
  211--252.

\bibitem{tarvainen2017mean}
{\sc Tarvainen, A., and Valpola, H.}
\newblock Mean teachers are better role models: Weight-averaged consistency
  targets improve semi-supervised deep learning results.
\newblock {\em arXiv preprint arXiv:1703.01780\/} (2017).

\bibitem{wang2020understanding}
{\sc Wang, T., and Isola, P.}
\newblock Understanding contrastive representation learning through alignment
  and uniformity on the hypersphere.
\newblock In {\em International Conference on Machine Learning\/} (2020),
  pp.~9929--9939.

\bibitem{wu2021federated}
{\sc Wu, Y., Zeng, D., Wang, Z., Shi, Y., and Hu, J.}
\newblock Federated contrastive learning for volumetric medical image
  segmentation.
\newblock In {\em International Conference on Medical Image Computing and
  Computer-Assisted Intervention\/} (2021), Springer, pp.~367--377.

\bibitem{xiao2020what}
{\sc Xiao, T., Wang, X., Efros, A.~A., and Darrell, T.}
\newblock What should not be contrastive in contrastive learning.
\newblock {\em CoRR abs/2008.05659\/} (2020).

\bibitem{yun2019cutmix}
{\sc Yun, S., Han, D., Oh, S.~J., Chun, S., Choe, J., and Yoo, Y.}
\newblock Cutmix: Regularization strategy to train strong classifiers with
  localizable features.
\newblock In {\em Proceedings of the IEEE/CVF International Conference on
  Computer Vision\/} (2019), pp.~6023--6032.

\bibitem{zhang2017mixup}
{\sc Zhang, H., Cisse, M., Dauphin, Y.~N., and Lopez-Paz, D.}
\newblock mixup: Beyond empirical risk minimization.
\newblock In {\em International Conference on Learning Representations\/}
  (2017).

\end{thebibliography}

\end{document}